%% file: main.tex
\documentclass[10pt,twocolumn,letterpaper]{article}

\usepackage[pagenumbers]{cvpr}
\usepackage[T1]{fontenc}
\usepackage{graphicx}
\usepackage{amsmath}
\usepackage{amssymb}
\usepackage{booktabs}
\usepackage{multirow}
\usepackage{bbm}
\usepackage{pifont}
\usepackage{enumitem}
\usepackage{algorithm}
\usepackage{listings}
\usepackage{amsfonts}
\usepackage{nicefrac}
\usepackage{microtype}
\usepackage{float}
\usepackage{caption}
\usepackage{wrapfig}
\usepackage{lipsum}
\usepackage[dvipsnames]{xcolor}
\usepackage[accsupp]{axessibility}
\usepackage{sidecap}

\usepackage{tabulary}
\newcolumntype{y}[1]{>{\raggedright\arraybackslash}p{#1pt}}
\newcolumntype{z}[1]{>{\raggedleft\arraybackslash}p{#1pt}}

\setlist[enumerate,1]{%
  label=\arabic*.,
}

\newlist{inlinelist}{enumerate*}{1}
\setlist*[inlinelist,1]{%
  label=(\roman*),
}

\definecolor{citecolor}{HTML}{0071bc}
\definecolor{bluegreen}{RGB}{117, 171, 188}
\newcommand{\gray}{\textcolor{gray}}

\usepackage[pagebackref,breaklinks,colorlinks, anchorcolor=blue, citecolor=citecolor]{hyperref}

\usepackage[capitalize]{cleveref}
\crefname{section}{Sec.}{Secs.}
\Crefname{section}{Section}{Sections}
\Crefname{table}{Table}{Tables}
\crefname{table}{Tab.}{Tabs.}

\newlength\savewidth
\newcommand{\pub}[1]{\color{gray}{\tiny{[{#1}]}}}

\def\cvprPaperID{****}
\def\confName{CVPR}
\def\confYear{2023}

\begin{document}

\title{Revisiting Weak-to-Strong Consistency\\ in Semi-Supervised Semantic Segmentation}

\author{Lihe~Yang$^1$~~~~~~~Lei~Qi$^2$~~~~~~~Litong~Feng$^3$~~~~~~~Wayne~Zhang$^3$~~~~~~~Yinghuan~Shi$^1$\thanks{Corresponding author.} \vspace{.5mm}\\
$^1$Nanjing University~~~~~~~$^2$Southeast University~~~~~~~$^3$SenseTime Research \\
\small{\url{https://github.com/LiheYoung/UniMatch}}
}

\maketitle

\input{section/abstract}
\input{section/introduction}
\input{section/related_work}
\input{section/method}
\input{section/experiment}
\input{section/conclusion}
\input{section/supp_singlecolumn}

\clearpage
{\small
\bibliographystyle{ieee_fullname}
\bibliography{egbib}
}

\end{document}

%% file: section/abstract.tex
\begin{abstract}
 In this work, we revisit the weak-to-strong consistency framework, popularized by FixMatch from semi-supervised classification, where the prediction of a weakly perturbed image serves as supervision for its strongly perturbed version. Intriguingly, we observe that such a simple pipeline already achieves competitive results against recent advanced works, when transferred to our segmentation scenario. Its success heavily relies on the manual design of strong data augmentations, however, which may be limited and inadequate to explore a broader perturbation space. Motivated by this, we propose an auxiliary feature perturbation stream as a supplement, leading to an expanded perturbation space. On the other, to sufficiently probe original image-level augmentations, we present a dual-stream perturbation technique, enabling two strong views to be simultaneously guided by a common weak view. Consequently, our overall Unified Dual-Stream Perturbations approach (UniMatch) surpasses all existing methods significantly across all evaluation protocols on the Pascal, Cityscapes, and COCO benchmarks. Its superiority is also demonstrated in remote sensing interpretation and medical image analysis. We hope our reproduced FixMatch and our results can inspire more future works.
\end{abstract}

%% file: section/introduction.tex
\section{Introduction}

Semantic segmentation aims to provide pixel-level predictions to images, which can be deemed as a dense classification task and is fundamental to real-world applications, \eg, autonomous driving. Nevertheless, conventional fully-supervised scenario \cite{long2015fully, zhao2017pyramid, zhang2018context} is extremely hungry for delicately labeled images by human annotators, greatly hindering its broad application to some fields where it is costly and even infeasible to annotate abundant images. Therefore, semi-supervised semantic segmentation \cite{souly2017semi} has been proposed and is attracting increasing attention. Generally, it wishes to alleviate the labor-intensive process via leveraging a large quantity of unlabeled images, accompanied by a handful of manually labeled images.

\begin{figure}
    \centering
    \includegraphics[width=0.82\linewidth]{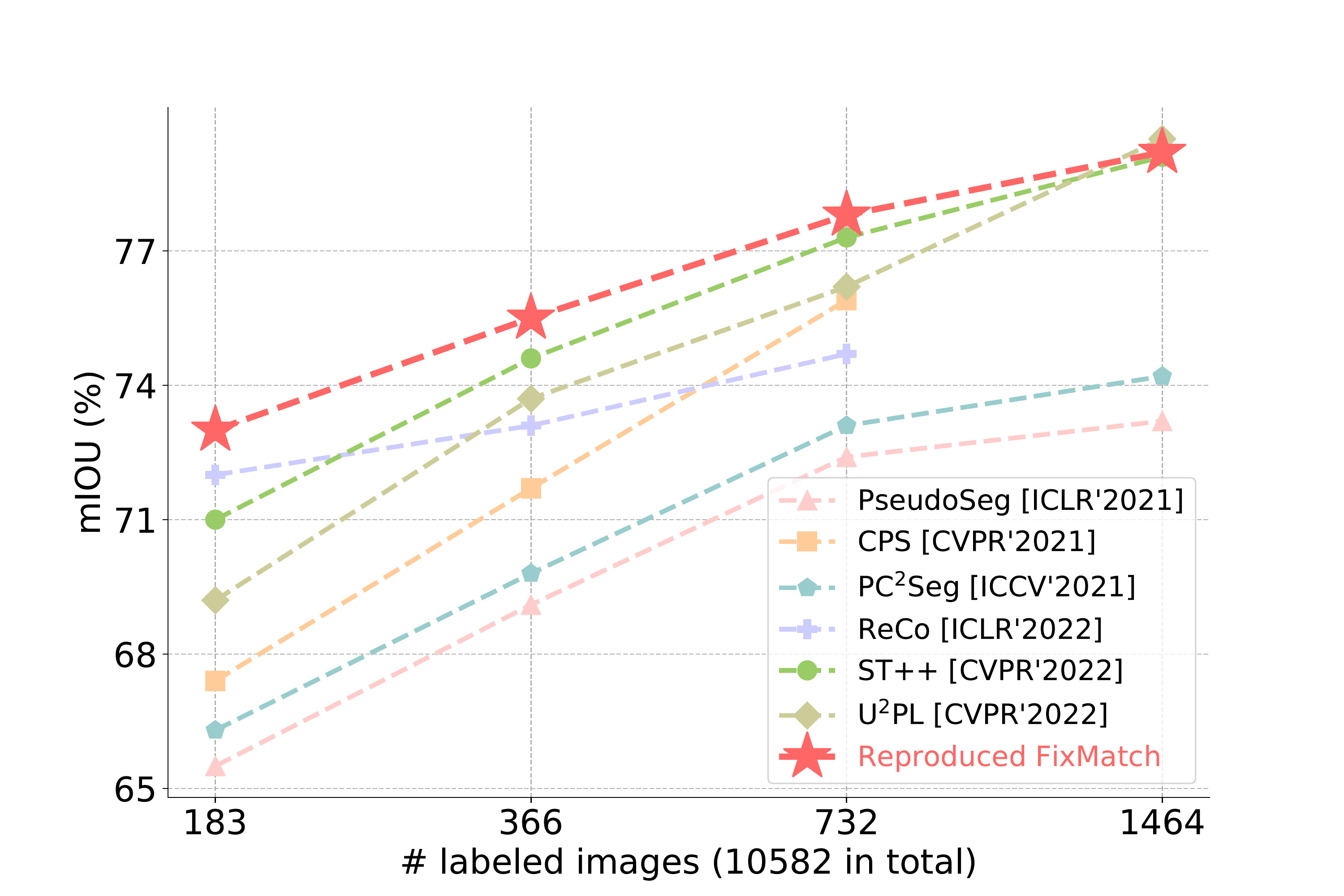}
    \vspace{-1mm}
    \caption{Comparison between state-of-the-art methods and our reproduced FixMatch \cite{sohn2020fixmatch} on the Pascal dataset.}
    \vspace{-4mm}
    \label{fig:fixmatch_sota}
\end{figure}

Following closely the research line of semi-supervised learning (SSL), advanced methods in semi-supervised semantic segmentation have evolved from GANs-based adversarial training paradigm \cite{goodfellow2014generative, souly2017semi, mittal2019semi} into the widely adopted consistency regularization framework \cite{french2019semi, ouali2020semi, zou2021pseudoseg, chen2021semi, hu2021semi, huo2020atso, wang2022semi} and reborn self-training pipeline \cite{yuan2021simple, he2021re, yang2022st++, guan2022unbiased}. In this work, we focus on the \emph{weak-to-strong} consistency regularization framework, which is popularized by FixMatch \cite{sohn2020fixmatch} from the field of semi-supervised classification, and then impacts many other relevant tasks \cite{liu2021unbiased, tang2021humble, xu2021end, melas2021pixmatch, xiao2022learning, xu2022cross}. The weak-to-strong approach supervises a strongly perturbed unlabeled image $x^s$ with the prediction yielded from its corresponding weakly perturbed version $x^w$, as illustrated in Figure~\ref{fig:fixmatch}. Intuitively, its success lies in that the model is more likely to produce high-quality prediction on $x^w$, while $x^s$ is more effective for our model to learn, since the strong perturbations introduce additional information as well as mitigate confirmation bias \cite{arazo2020pseudo}. We surprisingly notice that, so long as coupled with appropriate strong perturbations, FixMatch can indeed still exhibit powerful generalization capability in our scenario, obtaining superior results over state-of-the-art (SOTA) methods, as compared in Figure~\ref{fig:fixmatch_sota}. Thus, we select this simple yet effective framework as our baseline.

\begin{table}
\small
\centering
\setlength\tabcolsep{3mm}
    \begin{tabular}{r|ccccc}
    \toprule
    
    \multirow{2}{*}{Method} & \multicolumn{5}{c}{\# labeled images (10582 in total)} \\
    \cmidrule{2-6}
    ~ & 92 & 183 & 366 & 732 & 1464 \\
    \midrule
    w/o any SP & 39.5 & 52.7 & 65.5 & 69.2 & 74.6 \\
    
    w/ CutMix & \underline{56.7} & \underline{67.9} & \underline{71.9} & \underline{75.1} & \underline{78.3} \\
    
    w/ whole SP & \textbf{63.9} & \textbf{73.0} & \textbf{75.5} & \textbf{77.8} & \textbf{79.2} \\
    
    \bottomrule
    \end{tabular}
    \vspace{-2mm}
    \caption{The importance of image-level strong perturbations (SP) to FixMatch on the Pascal dataset. \emph{w/o any SP}: directly utilize hard label of $x^w$ to supervise its logits. \emph{w/ CutMix}: only use CutMix \cite{yun2019cutmix} as a perturbation. \emph{w/ whole SP}: strong perturbations contain color transformations from ST++ \cite{yang2022st++}, together with CutMix.}
    \vspace{-4mm}
    \label{tab:ablation_on_sp}
\end{table}

Through investigation of image-level strong perturbations, we observe that they play an indispensable role in making the FixMatch a rather strong competitor in semi-supervised semantic segmentation. As demonstrated in Table~\ref{tab:ablation_on_sp}, the performance gap between whether to adopt perturbations is extremely huge. Greatly inspired by these clues, we hope to inherit the spirit of strong perturbations from FixMatch, but also further strengthen them from two different perspectives and directions, namely \textit{expanding a broader perturbation space}, and \textit{sufficiently harvesting original perturbations}. Each of these two perspectives is detailed in the following two paragraphs respectively.

Image-level perturbations, \eg, color jitter and CutMix \cite{yun2019cutmix}, include heuristic biases, which actually introduce additional prior information into the bootstrapping paradigm of FixMatch, so as to capture the merits of consistency regularization. In case not equipped with these perturbations, FixMatch will be degenerated to a na\"ive online self-training pipeline, producing much worse results. Despite its effectiveness, these perturbations are totally constrained at the image level, hindering the model to explore a broader perturbation space and to maintain consistency at diverse levels. To this end, in order to expand original perturbation space, we design a unified perturbation framework for both raw images and extracted features. Concretely, on raw images, similar to FixMatch, pre-defined image-level strong perturbations are applied, while for extracted features of weakly perturbed images, an embarrassingly simple channel dropout is inserted. In this way, our model pursues the equivalence of predictions on unlabeled images at both the image and embedding level. These two perturbation levels can be complementary to each other. Distinguished from \cite{kuo2020featmatch, liu2022perturbed}, we separate different levels of perturbations into independent streams to avoid a single stream being excessively hard to learn.

On the other hand, current FixMatch framework merely utilizes a single strong view of each unlabeled image in a mini-batch, which is insufficient to fully exploit the manually pre-defined perturbation space. Considering this, we present a simple yet highly effective improvement to the input, where dual independent strong views are randomly sampled from the perturbation pool. They are then fed into the student model in parallel, and simultaneously supervised by their shared weak view. Such a minor modification even easily turns the FixMatch baseline into a stronger SOTA framework by itself. Intuitively, we conjecture that enforcing two strong views to be close to a common weak view can be regarded as minimizing the distance between these strong views. Hence, it shares the spirits and merits of contrastive learning \cite{chen2020simple, he2020momentum}, which can learn more discriminative representations and is proved to be particularly beneficial to our current task \cite{liu2022bootstrapping, wang2022semi}. We conduct comprehensive studies on the effectiveness of each proposed component. Our contributions can be summarized in four folds:
\begin{itemize}
\setlength\itemsep{-0.5mm}
    \item We notice that, coupled with appropriate image-level strong perturbations, FixMatch is still a powerful framework when transferred to the semantic segmentation scenario. A plainly reproduced FixMatch outperforms almost all existing methods in our current task.
    
    \item Built upon FixMatch, we propose a unified perturbation framework that unifies image-level and feature-level perturbations in independent streams, to exploit a broader perturbation space.
    
    \item We design a dual-stream perturbation strategy to fully probe pre-defined image-level perturbation space, as well as to harvest the merits of contrastive learning for discriminative representations.
    
    \item Our framework that integrates above two components, surpasses existing methods remarkably across all evaluation protocols on the Pascal, Cityscapes, and COCO. Notably, it also exhibits strong superiority in medical image analysis and remote sensing interpretation.
\end{itemize}

%% file: section/related_work.tex
\section{Related Work}

\noindent
\textbf{Semi-supervised learning (SSL).} The core issue in SSL lies in how to design reasonable and effective supervision signals for unlabeled data. Two main branches of methodology are proposed to tackle the issue, namely entropy minimization \cite{grandvalet2005semi, rosenberg2005semi, lee2013pseudo, xie2020self, zoph2020rethinking, pham2021meta} and consistency regularization \cite{laine2016temporal, sajjadi2016regularization, tarvainen2017mean, xie2019unsupervised, jeong2019consistency, berthelot2019mixmatch, berthelot2019remixmatch, miyato2018virtual, gong2021alphamatch, li2021comatch}. Entropy minimization, popularized by self-training \cite{lee2013pseudo}, works in a straightforward way via assigning pseudo labels to unlabeled data and then combining them with manually labeled data for further re-training. For another thing, consistency regularization holds the assumption that prediction of an unlabeled example should be invariant to different 
forms of perturbations. Among them, FixMatch \cite{sohn2020fixmatch} proposes to inject strong perturbations to unlabeled images and supervise training process with predictions from weakly perturbed ones to subsume the merits of both methodologies. Recently, FlexMatch \cite{zhang2021flexmatch} and FreeMatch \cite{wang2022freematch} consider learning status of different classes and then filter low-confidence labels with class-wise thresholds. Our method inherits from FixMatch, however, we investigate a more challenging and labor-intensive setting. More importantly, we demonstrate the significance of image-level strong perturbations, thereby managing to expand original perturbation space and take full advantage of pre-defined perturbations.

\noindent
\textbf{Semi-supervised semantic segmentation.} Earlier works \cite{mittal2019semi, souly2017semi} incorporate the GANs \cite{goodfellow2014generative} as an auxiliary supervision for unlabeled images via discriminating pseudo labels from manual labels. Motivated by the rapid progress in SSL, recent methods \cite{ouali2020semi, feng2020dmt, ke2020guided, mendel2020semi, zou2021pseudoseg, zhou2021c3, zhong2021pixel, alonso2021semi, zhang2021robust, liu2022bootstrapping, liu2022perturbed, kwon2022semi, zhang2022region} strive for simpler training paradigms from the perspective of consistency regularization and entropy minimization. During this trend, French \etal \cite{french2019semi} disclose Cutout \cite{devries2017improved} and CutMix \cite{yun2019cutmix} are critical to success of consistency regularization in segmentation. AEL \cite{hu2021semi} then designs an adaptive CutMix and sampling strategy to enhance the learning on under-performing classes. Inspired by contrastive learning, Lai \etal \cite{lai2021semi} propose to enforce predictions of the shared patch under different contextual crops to be same. And U$^2$PL \cite{wang2022semi} treats uncertain pixels as reliable negative samples to contrast against corresponding positive samples. Similar to the core spirit of co-training \cite{blum1998combining, qiao2018deep,zhang2018deep}, CPS \cite{chen2021semi} introduces dual models to supervise each other.

Other works from the research line of entropy minimization utilize a self-training pipeline to assign pseudo masks for unlabeled images in an offline manner. From this perspective, Yuan \etal \cite{yuan2021simple} claim excessive perturbations on unlabeled images are catastrophic to clean data distribution, and thus propose a separate batch normalization for these images. Concurrently, ST++ \cite{yang2022st++} points out that appropriate strong data perturbations are indeed extremely helpful to self-training. Moreover, to tackle the class bias issue encountered in pseudo labeling, He \etal \cite{he2021re} align class distributions between manual labels and pseudo labels. And USRN \cite{guan2022unbiased} clusters balanced subclass distributions as a regularization to alleviate the imbalance of pre-defined classes. 

To pursue elegance and efficacy, we adopt the weak-to-strong consistency regularization framework from FixMatch \cite{sohn2020fixmatch}. Our end-to-end baseline can be deemed as an improvement of \cite{french2019semi}, or a simplification of \cite{zou2021pseudoseg}. For instance, it strengthens image-level strong perturbations in \cite{french2019semi} with color transformations from \cite{yang2022st++}, and discards the calibration fusion module in \cite{zou2021pseudoseg}. With this neat but competitive baseline, we further probe a broader perturbation space, and fully exploit original image-level perturbations as well.

%% file: section/method.tex
\section{Method}

Algorithms in semi-supervised semantic segmentation aim to fully explore unlabeled images $\mathcal{D}^u = \{x^u_i\}$ with limited amount of annotations from labeled images $\mathcal{D}^l = \{(x^l_i, y^l_i)\}$. Our method is based on FixMatch \cite{sohn2020fixmatch}, so we first briefly review its core idea (\cref{sec:method_preliminary}). Following this, we introduce the two proposed components in detail, namely unified perturbations (\cref{sec:uniperb}), and dual-stream perturbations (\cref{sec:dusperb}). Finally, we summarize our overall Unified Dual-Stream Perturbations method (UniMatch) (\cref{sec:unimatch}).

\begin{figure}[t]
\centering
    \begin{minipage}{0.35\linewidth}
    \centering\captionsetup[subfigure]{justification=centering}
    \includegraphics[width=\linewidth]{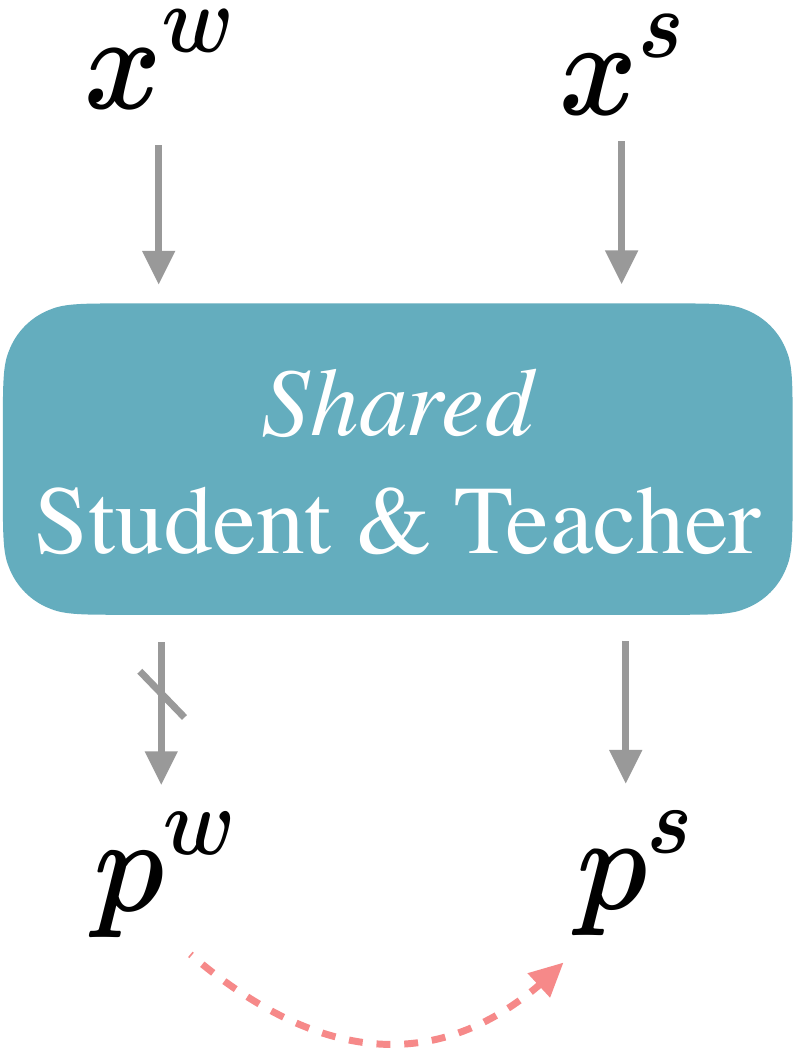}
    \subcaption{FixMatch}
    \label{fig:fixmatch}
    \end{minipage}
    \hspace{5mm}
    \begin{minipage}{0.45\linewidth}
    \includegraphics[width=\linewidth]{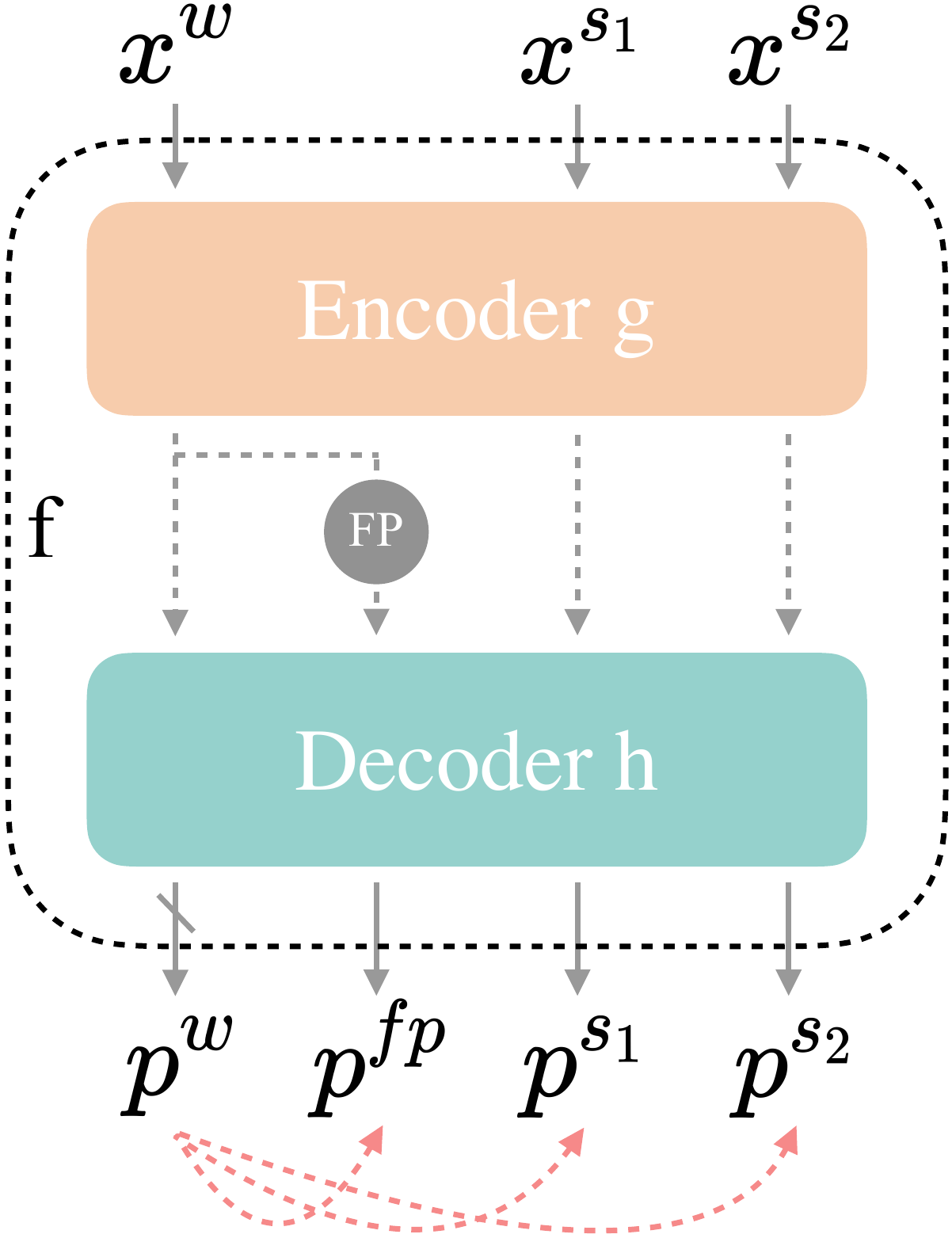}
    \subcaption{UniMatch}
    \label{fig:unimatch}
    \end{minipage}
\vspace{-1mm}
\caption{(a) The FixMatch baseline. (b) Our proposed unified dual-stream perturbations method (UniMatch). The \gray{FP} denotes feature perturbation, and the \textcolor{pink}{dashed curves} represent supervision.}
\vspace{-4mm}
\end{figure}

\subsection{\label{sec:method_preliminary}Preliminaries}

As aforementioned, FixMatch utilizes a weak-to-strong consistency regularization to leverage unlabeled data. Concretely, each unlabeled image $x^u$ is simultaneously perturbed by two operators, \ie, weak perturbation $\mathcal{A}^w$ such as cropping, and strong perturbation $\mathcal{A}^s$ such as color jitter. Then, the overall objective function is a combination of supervised loss $\mathcal{L}_s$ and unsupervised loss $\mathcal{L}_u$ as:
\begin{equation}
    \mathcal{L} = \frac{1}{2}(\mathcal{L}_s + \mathcal{L}_u).
\end{equation}
Typically, the supervised term $\mathcal{L}_s$ is the cross-entropy loss between model predictions and groundtruth labels. And the unsupervised loss $\mathcal{L}_u$ regularizes prediction of the sample under strong perturbations to be the same as that under weak perturbations, which can be formulated as:
\begin{equation}
    \mathcal{L}_u = \frac{1}{B_u}\sum \mathbbm{1}(\max(p^w) \geq \tau)\mathrm{H}(p^w, p^s),
    \label{eq:loss_u}
\end{equation}
where $B_u$ is the batch size for unlabeled data and $\tau$ is a pre-defined confidence threshold to filter noisy labels. $\mathrm{H}$ minimizes the entropy between two probability distributions:
\vspace{-1mm}
\begin{equation}
    p^w = \hat{F}(\mathcal{A}^w(x^u)); \hspace{2mm} p^s = F(\mathcal{A}^s(\mathcal{A}^w(x^u))),
\end{equation}
where the teacher model $\hat{F}$ produces pseudo labels on weakly perturbed images, while the student $F$ leverages strongly perturbed images for model optimization. In this work, we set $\hat{F}$ exactly the same as $F$ for simplicity, following FixMatch.

\subsection{\label{sec:uniperb}Unified Perturbations for Images and Features}

Apart from semi-supervised classification, the methodology in FixMatch has swept across a wide range of research topics and achieved booming success, such as semantic segmentation \cite{french2019semi, zou2021pseudoseg, hu2021semi}, object detection \cite{liu2021unbiased, tang2021humble, xu2021end}, unsupervised domain adaptation \cite{melas2021pixmatch}, and action recognition \cite{xiao2022learning, xu2022cross}. Despite its popularity, its efficacy actually heavily depends on delicately designed strong perturbations from researchers, whose optimal combinations and hyper-parameters are time-consuming to acquire. Besides, in some cases such as medical image analysis and remote sensing interpretation, it may require domain-specific knowledge to figure out promising ones. More importantly, they are completely constrained at the image level, hindering the student model to maintain multi-level consistency against more diverse perturbations.

\begin{figure}[t]
\centering
    \begin{minipage}{0.3\linewidth}
    \centering\captionsetup[subfigure]{justification=centering}
    \includegraphics[width=\linewidth]{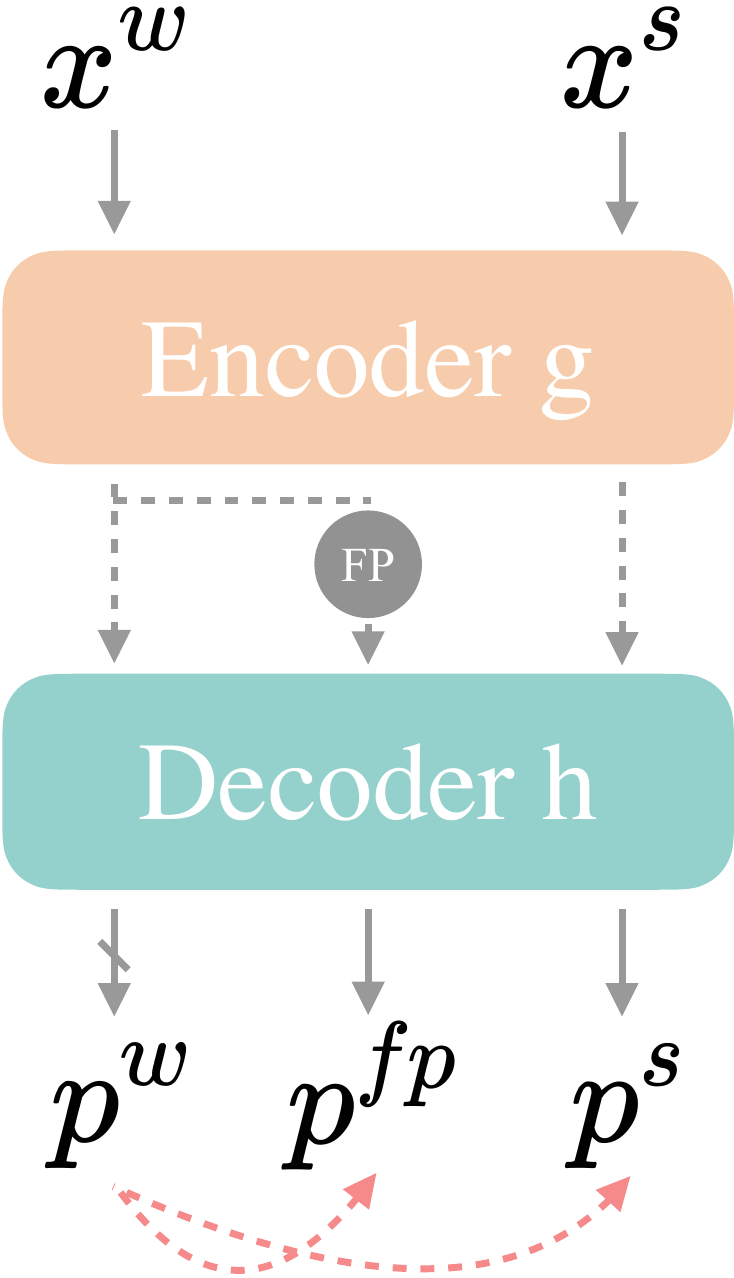}
    \subcaption{UniPerb}
    \label{fig:uniperb}
    \end{minipage}
    \hspace{1cm}
    \begin{minipage}{0.3\linewidth}
    \includegraphics[width=\linewidth]{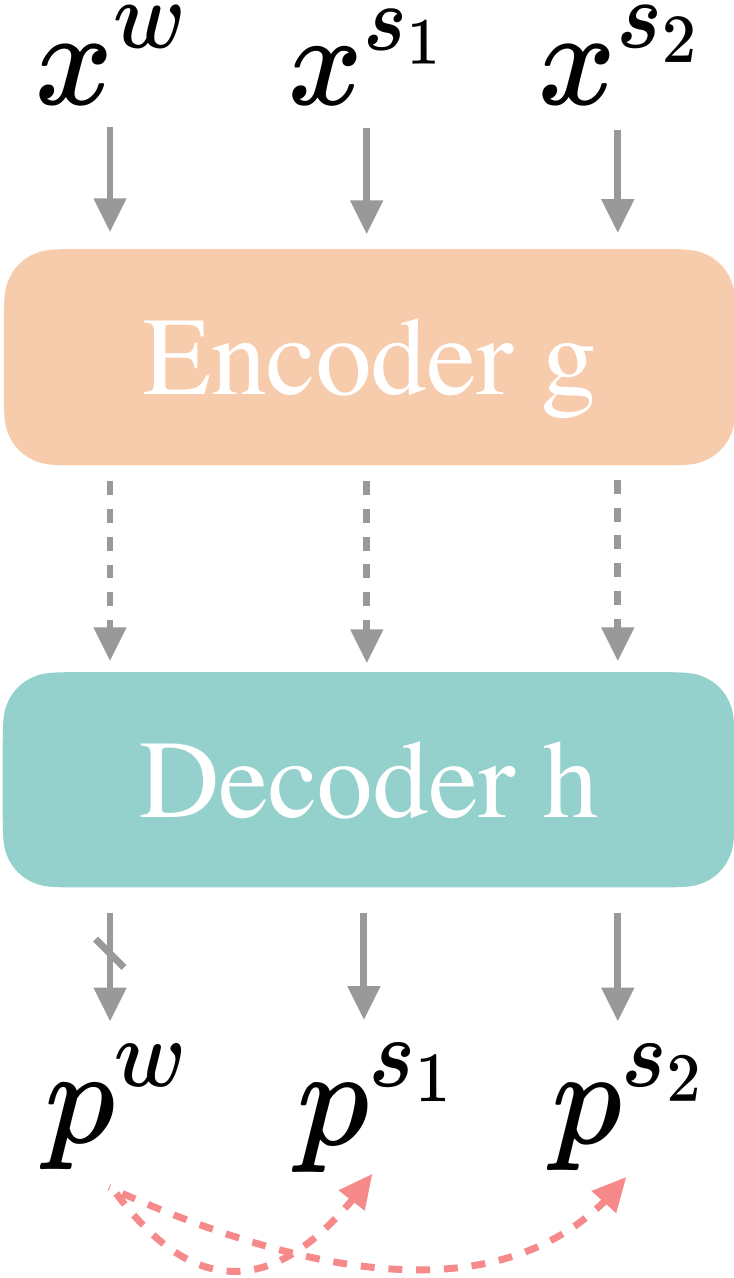}
    \subcaption{DusPerb}
    \label{fig:dusperb}
    \end{minipage}
\vspace{-2mm}
\caption{The two sub-frameworks, \ie, UniPerb and DusPerb, that are integrated into our final UniMatch framework. The \gray{FP} denotes feature perturbation, which is a simple Dropout in this work.}
\vspace{-4mm}
\end{figure}

To this end, in order to construct a broader perturbation space, built on top of FixMatch, we propose to inject perturbations on features of the \emph{weakly} perturbed image $x^w$. We choose to separate different levels of perturbations into multiple independent feedforward streams, enabling the student to achieve targeted consistency in each stream more directly. Formally, a segmentation model $f$ can be decomposed into an encoder $g$ and a decoder $h$. In addition to acquired $p^w$ and $p^s$ in FixMatch, we also obtain $p^{fp}$ from an auxiliary feature perturbation stream by:
\begin{align}
e^w &= g(x^w),\\
p^{fp} &= h(\mathcal{P}(e^w)),
\end{align}
where $e^w$ is extracted features of $x^w$, and $\mathcal{P}$ denotes feature perturbations, \eg, dropout or adding uniform noise.

\begin{algorithm}[t]
\caption{\small{Pseudocode of UniMatch in a PyTorch-like style.}}
\label{alg:unimatch}
\definecolor{codeblue}{rgb}{0.25,0.5,0.5}
\lstset{
  backgroundcolor=\color{white},
  basicstyle=\fontsize{7pt}{7pt}\ttfamily\selectfont,
  columns=fullflexible,
  breaklines=true,
  captionpos=b,
  commentstyle=\fontsize{7pt}{7pt}\color{codeblue},
  keywordstyle=\fontsize{7pt}{7pt}
}
\begin{lstlisting}[language=python]
# f: network, composed of an encoder g and a decoder h
# aug_w/aug_s: weak/strong image-level perturbations

for x in loader_u:
    # one weak view and two strong views as input
    x_w = aug_w(x)
    x_s1, x_s2 = aug_s(x_w), aug_s(x_w)
    
    # feature of weakly perturbed image
    feat_w = g(x_w)
    # perturbed feature
    feat_fp = nn.Dropout2d(0.5)(feat_w)
    
    # four predictions from four forward streams
    p_w, p_fp = h(torch.cat((feat_w, feat_fp))).chunk(2)
    p_s1, p_s2 = f(torch.cat((x_s1, x_s2))).chunk(2)
    
    # hard (one-hot) pseudo mask
    mask_w = p_w.argmax(dim=1).detach()
    
    # loss from image- and feature-level perturbation
    criterion = nn.CrossEntropyLoss()
    p_s = torch.cat((p_s1, p_s2))
    loss_s = criterion(p_s, mask_w.repeat(2, 1, 1))
    loss_fp = criterion(p_fp, mask_w)
    
    # final unsupervised loss
    loss_u = (loss_s + loss_fp) / 2.0
\end{lstlisting}
\end{algorithm}

Overall, as exhibited in Figure~\ref{fig:uniperb}, three feedforward streams are maintained for each unlabeled mini-batch, which are 
\begin{inlinelist}
\item the simplest stream: $x^w \rightarrow f \rightarrow p^w$,
\item image-level strong perturbation stream: $x^s \rightarrow f \rightarrow p^s$, and
\item our introduced feature perturbation stream: $x^w \rightarrow g \rightarrow \mathcal{P} \rightarrow h \rightarrow p^{fp}$.
\end{inlinelist}
In this way, the student model is enforced to be consistent to unified perturbations at both image and feature level. We name it as UniPerb for convenience. The unsupervised loss $\mathcal{L}_u$ is formulated as:
\begin{equation}
\small
    \mathcal{L}_u = \frac{1}{B_u}\sum \mathbbm{1}(\max(p^w) \geq \tau)\big(\mathrm{H}(p^w, p^s) + \mathrm{H}(p^w, p^{fp})\big).
\end{equation}
It should be noted that, we do not aim at proposing a novel feature perturbation approach in this work. Actually, an embarrassingly simple channel dropout (\texttt{nn.Dropout2d} in PyTorch) is well-performed enough. Furthermore, distinguished from recent work \cite{liu2022perturbed} that mixes three levels of perturbations into a single stream, we highlight the necessity of separating perturbations of different properties into independent streams, which is evidenced in our ablation studies. We believe that, image-level perturbations can be well complemented by feature-level perturbations.

\subsection{\label{sec:dusperb}Dual-Stream Perturbations}

Motivated by the tremendous advantages of image-level strong perturbations, we wish to fully explore them. We are inspired by recent progress in self-supervised learning and semi-supervised classification, that constructing multiple views for unlabeled data as inputs can better leverage the perturbations. For instance, SwAV \cite{caron2020unsupervised} proposes a novel technique called \texttt{multi-crop}, enforcing the local-to-global consistency among a bag of views of different resolutions. Likewise, ReMixMatch \cite{berthelot2019remixmatch} produces multiple strongly augmented versions for the model to learn. 

Therefore, we wonder whether such a simple idea can also benefit our semi-supervised semantic segmentation. We make a straightforward attempt that, rather than feeding a single $p^s$ into the model, we independently yield dual-stream perturbations  ($x^{s_1}$, $x^{s_2}$) from $x^w$ by strong perturbation pool $\mathcal{A}^s$. Since $\mathcal{A}^s$ is pre-defined but non-deterministic, $x^{s_1}$ and $x^{s_2}$ are not equal. This dual-stream perturbation framework (DusPerb) is displayed in Figure~\ref{fig:dusperb}. 

Intriguingly, such a minor modification brings consistent and substantial improvements over original FixMatch under all partition protocols in our segmentation scenario, establishing new state-of-the-art results. It is validated in our ablation studies that, the performance gain is non-trivial, not credited to a doubled unlabeled batch size. We conjecture that regularizing two strong views with a shared weak view can be regarded as enforcing consistency between these two strong views as well. Intuitively, suppose $k_w$ is the classifier weight of the class predicted by $x^w$, and ($q_{s_1}$, $q_{s_2}$) are features of images ($x^{s_1}$, $x^{s_2}$), then in our adopted cross entropy loss, we maximize $q_j\cdot k_w$ against $\sum_{i=0}^C q_j\cdot k_i$, where $j \in \{s_1, s_2\}$, and $k_i$ is classifier weight of class $i$. It thus can be deemed that we are also maximizing the similarity between $q_{s_1}$ and $q_{s_2}$. So the InfoNCE loss \cite{van2018representation} is satisfied:
\begin{equation}
\small
    \mathcal{L}_{s_1\leftrightarrow s_2} = -\log\frac{\exp(q_{s_1}\cdot q_{s_2} )}{\sum_{i=0}^{C}\exp(q_j \cdot k_i)}, ~~s.t., j \in \{s_1, s_2\},
\end{equation}
where $q_{s_1}$ and $q_{s_2}$ are positive pairs, while all other classifier weights except $k_w$ are negative samples.

Hence, it shares the spirits of contrastive learning \cite{chen2020simple, he2020momentum, chen2021exploring}, which is able to learn discriminative representations and has been proved to be highly meaningful to our task \cite{liu2022bootstrapping, wang2022semi}.

\input{table/pascal_origin}

\subsection{\label{sec:unimatch}Our Holistic Framework: UniMatch}

To sum up, we present two key techniques to leverage unlabeled images, namely UniPerb and DusPerb. Our holistic framework (dubbed as UniMatch) that integrates both approaches is illustrated in Figure~\ref{fig:unimatch}. The corresponding pseudocode is provided in Algorithm~\ref{alg:unimatch}. In comparison with FixMatch, two auxiliary feedforward streams are maintained, one for perturbation on features of $x^w$, and the other for multi-view learning of ($x^{s_1}$, $x^{s_2}$). The final unsupervised term is computed as:
\vspace{-2mm}
\begin{multline}
    \mathcal{L}_u = \frac{1}{B_u}\sum \mathbbm{1}(\max(p^w) \geq \tau)\cdot \\
    \big(\lambda\mathrm{H}(p^w, p^{fp}) + \frac{\mu}{2}\big(\mathrm{H}(p^w, p^{s_1}) + \mathrm{H}(p^w, p^{s_2})\big)\big).
\end{multline}
It is clarified that feature-level and image-level perturbation streams have their own properties and advantages, thus their loss weights $\lambda$ and $\mu$ are equally set as 0.5. The $\textrm{H}$ in $\mathcal{L}_u$ is a regular cross-entropy loss. The confidence threshold $\tau$ is set as 0.95 for all datasets except Cityscapes, where $\tau$ is 0.

%% file: table/pascal_origin.tex
\begin{table}[t]
	\centering
	\small
	
	\begin{tabular}{z{44}y{28}|ccccc}
		\toprule
		\multicolumn{2}{c|}{\textbf{Pascal}} & 92 & 183 & 366 & 732 & 1464 \\
  
        \midrule
  
        SupBaseline\!\!&\!\!
		& 44.0 & 52.3 & 61.7 & 66.7 & 72.9 \\

        PC$^2$Seg \cite{zhong2021pixel}\!\!&\!\!\!\!\pub{ICCV'21} & 56.9 & 64.6 & 67.6 & 70.9 & 72.3 \\
        
        \midrule
        
        UniMatch\!\!&\hspace{-2.4mm}| \textbf{\gray{RN-50}} & \textbf{71.9} & \textbf{72.5} & \textbf{76.0} & \textbf{77.4} & \textbf{78.7} \\
        
        \midrule
		\midrule
		
		SupBaseline\!\!&\!\!
		& 45.1 & 55.3 & 64.8 & 69.7 & 73.5 \\
		
		CPS \cite{chen2021semi}\!\!&\!\!\!\!\pub{CVPR'21} & 64.1 & 67.4 & 71.7 & 75.9 & - \\
		
		ST++ \cite{yang2022st++}\!\!&\!\!\!\!\pub{CVPR'22} & 65.2 & 71.0 & 74.6 & 77.3 & 79.1 \\
		
		U$^2$PL \cite{wang2022semi}\!\!&\!\!\!\!\pub{CVPR'22} & 68.0 & 69.2 & 73.7 & 76.2 & 79.5\\
		
		PS-MT \cite{liu2022perturbed}\!\!&\!\!\!\!\pub{CVPR'22} & 65.8 & 69.6 & 76.6 & 78.4 & 80.0 \\
		
		PCR \cite{pcr}\!\!&\!\!\!\!\pub{NeurIPS'22} & 70.1 & 74.7 & 77.2 & 78.5 & 80.7 \\

        \midrule
        
		UniMatch\!\!&\hspace{-2.4mm}| \textbf{\gray{RN-101}} & \textbf{75.2} & \textbf{77.2} & \textbf{78.8} & \textbf{79.9} & \textbf{81.2} \\
		
		\bottomrule
	\end{tabular}
	\vspace{-2mm}
	\caption{Comparison with SOTAs on the \textbf{Pascal}. Labeled images are from the \emph{original high-quality} training set. The integers (\eg, 92) in the head denote the number of labeled images. Except ST++, the training resolution of other works is larger than us: 512 \textit{vs.}~321.}
	\vspace{-2mm}
	\label{table:pascal_original}
\end{table}

%% file: section/experiment.tex
\section{Experiments}

\input{table/pascal_all}

\subsection{\label{sec:implement_details}Implementation Details}

For a fair comparison with prior works, we mainly adopt DeepLabv3+ \cite{chen2018encoder} based on ResNet \cite{he2016deep} as our segmentation model. The ResNet uses an output stride of 16 across all experiments to save memory and speed up training. During training, each mini-batch is composed of 8 labeled images and 8 unlabeled images. The initial learning rate is set as 0.001, 0.005, and 0.004 for Pascal, Cityscapes, and COCO respectively, with a SGD optimizer. The model is trained for 80, 240, and 30 epochs under a poly learning rate scheduler. We assemble the color transformations from ST++ \cite{yang2022st++} and CutMix \cite{yun2019cutmix} to form our $\mathcal{A}^s$. A raw image is resized between 0.5 and 2.0, cropped, and flipped to obtain its weakly augmented version $x^w$. The training resolution is set as 321, 801, and 513 for these three datasets. By default, we adopt a channel dropout of 50\% probability (\texttt{nn.Dropout2d(0.5)} in PyTorch) as our feature perturbation, which is inserted at the intersection of the encoder and decoder.

\input{table/cityscapes}
\input{table/coco}

\vspace{-1mm}
\subsection{\label{sec:compare_sota}Comparison with State-of-the-Art Methods}

\noindent
\textbf{Pascal VOC 2012.} The Pascal dataset \cite{everingham2015pascal} is originally constructed of 1464 high-quality training images. Later, it is expanded by extra coarse annotations from the SBD \cite{hariharan2011semantic}, resulting in 10582 training images. There are three protocols to select labeled images: (1) \textbf{(the most convincing one)} select from the original high-quality training images, (2) select from the blended 10582 training images, and (3) prioritize the high-quality set, and if not enough, select from the expanded set. Under the first protocol (Table \ref{table:pascal_original}), our UniMatch outperforms existing methods tremendously. We even adopt a smaller training size of 321 than most recent works of 512. In addition, for the other two protocols (Table \ref{table:pascal_all}), we train UniMatch at two resolutions of 321 and 513. It still gains remarkable improvements over prior works.

\noindent
\textbf{Cityscapes.} This dataset \cite{cordts2016cityscapes} focuses on urban scenes, consisting of 2975 high-resolution training images and 500 validation images. We follow two techniques in previous SOTA works \cite{chen2021semi, hu2021semi, wang2022semi, liu2022perturbed, pcr}, \ie, sliding window evaluation and online hard example mining (OHEM) loss. Results under ResNet-50 and ResNet-101 are reported in Table~\ref{table:cityscapes}. Our results are consistently much better than existing best ones.

\input{table/ablation_uniperb_dusperb}

\input{table/citysapes_unimatch_fixmatch}
\input{table/coco_unimatch_fixmatch}

\noindent
\textbf{MS COCO.} The COCO dataset \cite{lin2014microsoft}, composed of 118k/5k training/validation images, is a quite challenging benchmark, containing 81 classes to predict. Therefore, it was rarely explored before in semi-supervised works of segmentation. However, in view of the seemingly saturate performance on the Pascal and Cityscapes, we believe it will be more practical to evaluate our algorithms on this dataset. We adopt exactly the same setting and backbone (Xception-65 \cite{chollet2017xception}) as PseudoSeg \cite{zou2021pseudoseg}. As evidenced by Table~\ref{table:coco}, our UniMatch significantly surpasses all available methods by 1.4\%-4.5\%.

\subsection{\label{sec:ablation}Ablation Studies}

Unless otherwise specified, we mainly conduct ablation studies on the Pascal dataset extensively with ResNet-101.

\noindent
\textbf{Improvement over the FixMatch baseline.} We conduct this most important ablation in Table \ref{table:ablation_uniperb_dusperb}, \ref{table:citys_unimatch_fixmatch}, and \ref{table:coco_unimatch_fixmatch} for all the three benchmarks respectively. It is clear that our UniMatch consistently improves the strong baseline by a large margin.

\noindent
\textbf{Individual effectiveness of UniPerb and DusPerb.} In Table~\ref{table:ablation_uniperb_dusperb}, we first demonstrate that our reproduced FixMatch is a strong competitor against previous SOTA methods. Then built upon FixMatch, both UniPerb and DusPerb facilitate this baseline by a large margin. Lastly, our overall UniMatch that integrates both components achieves the best results.

\noindent
\textbf{The improvement of diverse perturbations is non-trivial.} Our UniMatch utilizes three views, \ie, one feature perturbation view and dual image perturbation views. We wish to validate that constructing diverse perturbations is beneficial, much better than blindly maintaining three parallel image perturbations. So we design a simple counterpart that adopts three image-level strong perturbation views. As displayed in Table~\ref{table:ablation_3views}, our UniMatch is consistently superior to it, indicating the improvement brought by UniMatch is not credited to blindly increasing views, but the diversity counts.

\begin{figure}[t]
    \centering
    \includegraphics[width=0.75\linewidth]{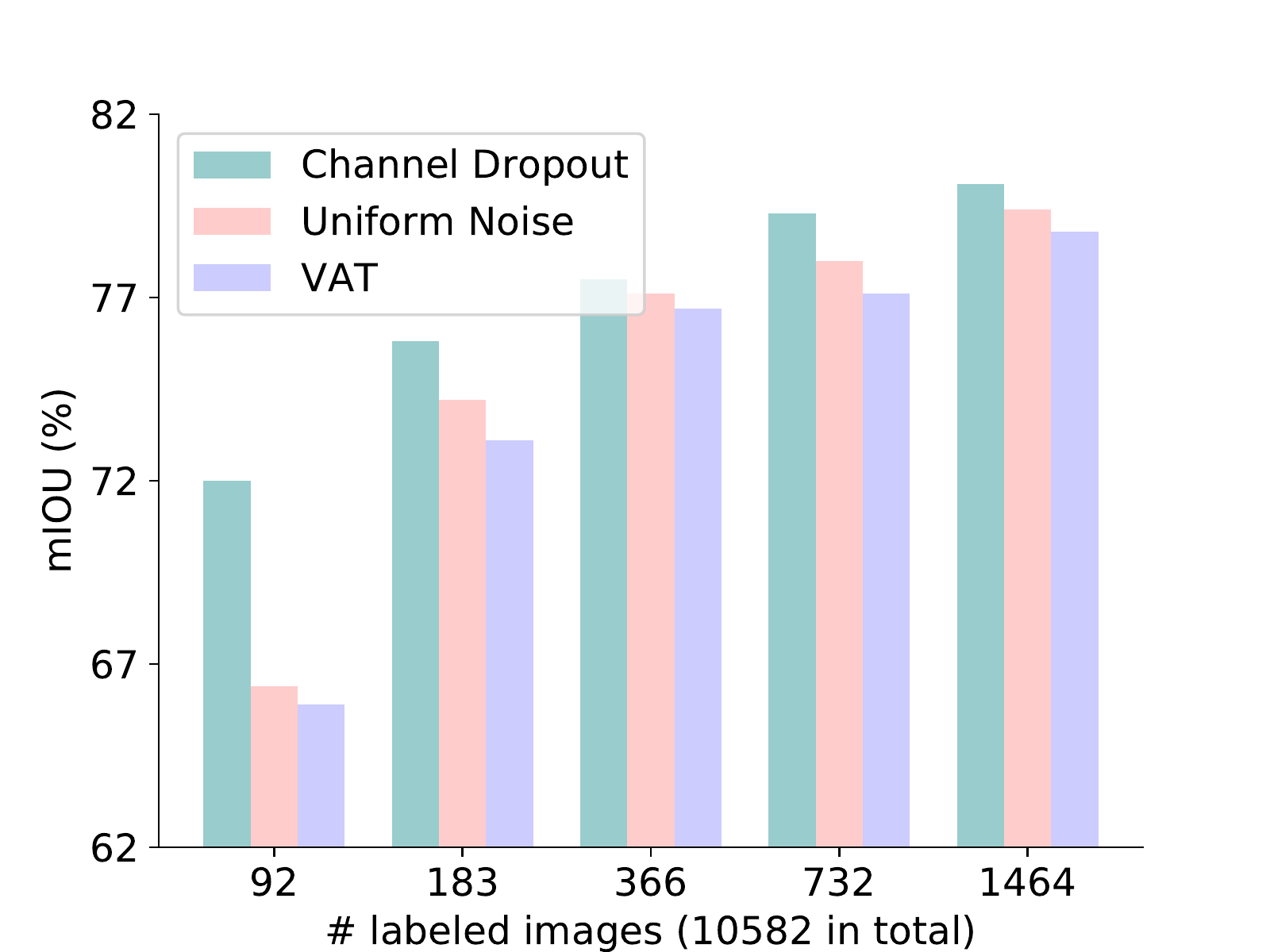}
    \vspace{-1mm}
    \caption{Ablation study on the efficacy of various feature perturbation strategies in our UniPerb method.}
    \vspace{-2mm}
    \label{fig:fp_variants}
\end{figure}

\input{table/ablation_3views}
\input{table/ablation_dusperb_2xbs}

\noindent
\textbf{The improvement of dual-stream perturbation is non-trivial.} It might have been noticed that in our DusPerb, the number of unlabeled images in each mini-batch is doubled, since each unlabeled image is strongly perturbed twice. Hence, it might be argued that the improvement is due to larger batch size. Considering this concern, we further carry out an ablation study on the FixMatch with a twice larger batch size (keep the same training iterations) or a twice longer training epochs. As presented in Table~\ref{table:ablation_dusperb_2xbs}, although increasing the unlabeled batch size or lengthening training epochs improves the FixMatch baseline in most cases, they are still evidently lagging behind our DusPerb.

\noindent
\textbf{The necessity of separating image- and feature-level perturbations into independent streams.} PS-MT \cite{liu2022perturbed} mixes three levels of perturbations into a single feedforward stream, however, we claim that separating perturbations with different properties into independent streams will enable the model to achieve the targeted invariance more directly, and also avoid a single stream being excessively hard. To confirm it, we inject the dropout on the features of \emph{strongly} perturbed images, forming a stream of hybrid view. As shown is  Table~\ref{table:ablation_shared_stream}, one hybrid view is inferior to one image perturbation view. Moreover, we attempt to adopt two hybrid views, but it is still worse than our separate practice in UniPerb.

\begin{figure}
    \centering
    \includegraphics[width=0.75\linewidth]{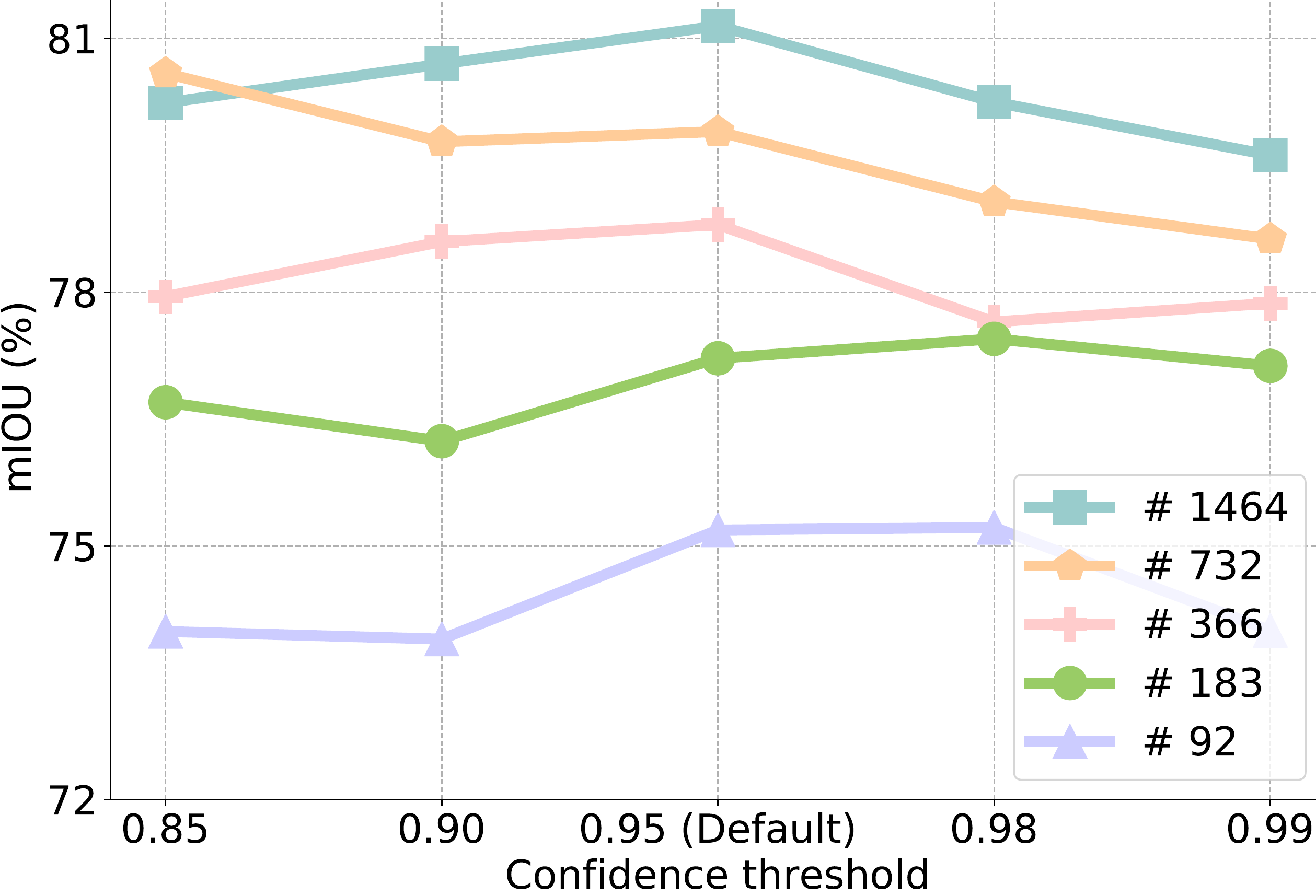}
    \vspace{-1mm}
    \caption{Ablation study on different values of confidence threshold in our UniMatch method on the \textbf{Pascal} dataset. Numbers in the legend denote the number of labeled images.}
    \vspace{-2mm}
    \label{fig:conf_thresh}
\end{figure}

\input{table/ablation_shared_stream}

\input{table/ablation_num_of_views}

\input{table/cd_whu_levir}

\noindent
\textbf{More perturbation streams.} We also attempt to increase the number of image- and feature-level perturbation streams in Table \ref{table:ablation_num_of_views}. It is observed that, increasing the perturbation streams does not necessarily result in higher performance. This also indicates that, the two image streams and one feature stream in our UniMatch are well-performed enough.

\noindent
\textbf{Other feature perturbation strategies.} We use a simplest form of feature perturbation in our method, which is a channel dropout. There are some other options available, such as uniform noise and virtual adversarial training (VAT) \cite{miyato2018virtual}. We follow \cite{ouali2020semi} to set the hyper-parameters in these strategies. And all these options are compared in Figure~\ref{fig:fp_variants}. It can be concluded that a channel dropout performs best.

\begin{figure}[t]
\centering
    \includegraphics[width=0.8\linewidth]{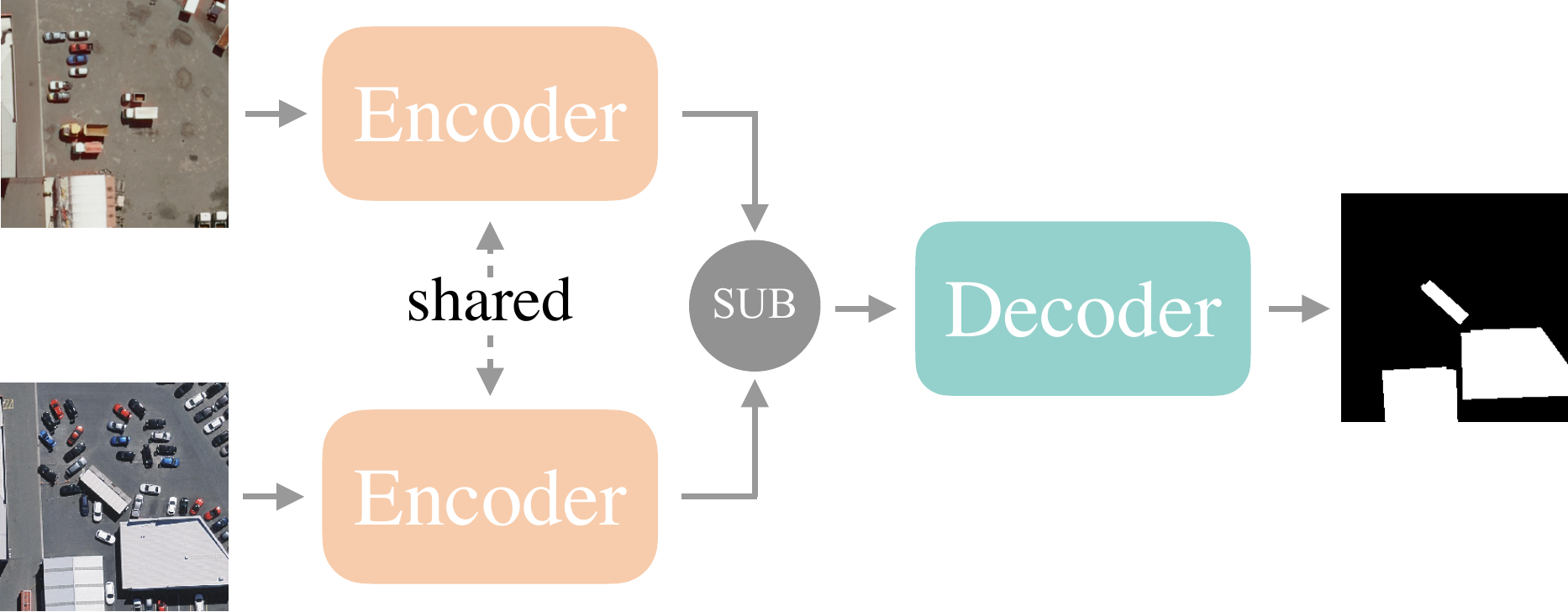}
    \vspace{-1mm}
    \caption{A typical framework in change detection task. Features extracted by the encoder are subtracted to be fed into the decoder.}
    \vspace{-2mm}
    \label{fig:cd_pipeline}
\end{figure}

\input{table/ablation_fp_location}

\noindent
\textbf{Value of the confidence threshold $\tau$.} We ablate this hyper-parameter on the Pascal in Figure~\ref{fig:conf_thresh}. It is observed that $\tau$ of 0.95 works best for the Pascal.

\noindent
\textbf{Locations to insert feature perturbations.} Our feature perturbations are injected at the intersection of the encoder and decoder. Previous work \cite{ouali2020semi} also performs perturbations to the input of final classifier. We compare the two locations in Table~\ref{tab:ablation_fp_location}. It is observed that our practice is much better.

\subsection{Application to More Segmentation Scenarios\label{sec:more_application}}

We have validated our UniMatch in common benchmarks of natural images. Here, we further carry out extra experiments in two highly practical and critical scenarios, \ie, remote sensing interpretation and medical image analysis. In both scenarios, unlabeled data is easy and cheap to acquire, while manual annotations are extremely expensive.

\noindent
\textbf{Remote Sensing Interpretation.} We focus on the change detection task in this scenario, due to its wide application demand and strict labeling requirement. Given a pair of bi-temporal images, \ie, two images for the same region but of different times, the changed regions are required to be highlighted. It can be simply deemed as a binary segmentation problem. A typical framework is illustrated in Figure~\ref{fig:cd_pipeline}. Following the latest work SemiCD \cite{bandara2022revisiting}, we validate our UniMatch on two popular benchmarks, \ie, WHU-CD \cite{ji2018fully} and LEVIR-CD \cite{chen2020spatial}. We attempt on two networks, \ie PSPNet and DeepLabv3+, both based on ResNet-50. As shown in Table~\ref{tab:whu_levir_cd}, UniMatch outperforms SemiCD \cite{bandara2022revisiting} impressively.

\input{table/acdc}

\noindent
\textbf{Medical Image Analysis.} We follow a recent work \cite{luo2021semi} to investigate semi-supervised medical image segmentation on the ACDC dataset \cite{bernard2018deep}. As shown in Table~\ref{tab:acdc}, our UniMatch improves the SOTAs significantly, \eg, +23.3\% given 3 labeled cases. Our result of mere 1 labeled case even surpasses others with 3 cases, and is on par with others using 7 cases.

For implementation details of these two scenarios, please refer to our open-sourced code.
    

%% file: table/pascal_all.tex
\begin{table}[t]
	\centering
	\small
	
	\begin{tabular}{r|ccc|ccc}
		\toprule
		\multirow{2}{*}{\textbf{Pascal}} & \multicolumn{3}{c|}{ResNet-50} & \multicolumn{3}{c}{ResNet-101} \\
		
		\cmidrule{2-7}
		~ & 1/16 & 1/8 & 1/4 & 1/16 & 1/8 & 1/4 \\
		
		\midrule
		
		SupBaseline & 61.2 & 67.3 & 70.8 & 65.6 & 70.4 & 72.8 \\
		
		CAC \cite{lai2021semi} & 70.1 & 72.4 & 74.0 & 72.4 & 74.6 & 76.3 \\
		
		ST++ \cite{yang2022st++} & 72.6 & 74.4 & 75.4 & 74.5 & 76.3 & 76.6 \\
		
		\midrule
		
		UniMatch | \textbf{\gray{321}} & \textbf{74.5} & \textbf{75.8} & \textbf{76.1} & \textbf{76.5} & \textbf{77.0} & \textbf{77.2} \\
		
		\midrule
		\midrule
		
		SupBaseline & 62.4 & 68.2 & 72.3 & 67.5 & 71.1 & 74.2 \\
		
		CPS \cite{chen2021semi} & 72.0 & 73.7 & 74.9 & 74.5 & 76.4 & 77.7 \\
	    
		U$^2$PL \cite{wang2022semi} & 72.0 & 75.1 & 76.2 & 74.4 & 77.6 & 78.7 \\
		
		PS-MT \cite{liu2022perturbed} & 72.8 & 75.7 & 76.4 & 75.5 & 78.2 & 78.7 \\
		
		\midrule
		
		UniMatch | \textbf{\gray{513}} & \textbf{75.8} & \textbf{76.9} & \textbf{76.8} & \textbf{78.1} & \textbf{78.4} &  \textbf{79.2} \\

        \midrule
        \midrule

        SupBaseline$^\dag$ & 67.7 & 71.9 & 74.5 & 70.6 & 75.0 & 76.5 \\

        U$^2$PL$^\dag$ \cite{wang2022semi} & 74.7 & 77.4 & 77.5 & 77.2 & 79.0 & 79.3 \\

        UniMatch$^\dag$  | \textbf{\gray{513}} & \textbf{78.1} & \textbf{79.0} & \textbf{79.1} & \textbf{80.9} & \textbf{81.9} & \textbf{80.4} \\
        
		\bottomrule
	\end{tabular}
	\vspace{-2mm}
	\caption{Comparison with SOTAs on the \textbf{Pascal}. Labeled images are sampled from the \emph{blended} training set. The \textbf{\gray{321}} and \textbf{\gray{513}} denote the training resolution. The fractions in the head denote the proportion of labeled images. We reproduce the RN-50 results of U$^2$PL. $\dag$: Prioritizing selecting labeled images from the high-quality set.}
	\vspace{-4mm}
	\label{table:pascal_all}
\end{table}

%% file: table/cityscapes.tex
\begin{table}[t]
\setlength\tabcolsep{2.4mm}
    \centering
    \small
    \begin{tabular}{z{45}y{28}|ccccc}
    \toprule
    
    \multicolumn{2}{c|}{\multirow{2}{*}{\textbf{Cityscapes}}} & 1/16 & 1/8 & 1/4 & 1/2 \\
    
    ~ & ~ & (186) & (372) & (744) & (1488) \\
    
    \midrule
    
    SupBaseline\!\!&\!\! & 63.3 & 70.2 & 73.1 & 76.6 \\
    
    PS-MT \cite{liu2022perturbed}\!\!&\!\!\!\!\pub{CVPR'22} & - & 75.8 & 76.9 & 77.6 \\
    
    U$^2$PL$^\ddag$ \cite{wang2022semi}\!\!&\!\!\!\!\pub{CVPR'22} & 70.6 & 73.0 & 76.3 & 77.2 \\
    
    \midrule
    
    UniMatch\!\!&\hspace{-2.4mm}| \textbf{\gray{RN-50}} & \textbf{75.0} & \textbf{76.8} & \textbf{77.5} & \textbf{78.6} \\
    
    \midrule
    \midrule
    
    SupBaseline\!\!&\!\! & 66.3 & 72.8 & 75.0 & 78.0 \\
    
    CPS \cite{chen2021semi}\!\!&\!\!\!\!\pub{CVPR'21} & 69.8 & 74.3 & 74.6 & 76.8 \\
		
    AEL \cite{hu2021semi}\!\!&\!\!\!\!\pub{NeurIPS'21} & 74.5 & 75.6 & 77.5 & 79.0 \\
	
    PS-MT \cite{liu2022perturbed}\!\!&\!\!\!\!\pub{CVPR'22} & - & 76.9 & 77.6 & 79.1 \\
	
    U$^2$PL \cite{wang2022semi}\!\!&\!\!\!\!\pub{CVPR'22} & 74.9 & 76.5 & 78.5 & 79.1 \\
	
    PCR \cite{pcr}\!\!&\!\!\!\!\pub{NeurIPS'22} & 73.4 & 76.3 & 78.4 & 79.1 \\
	
    \midrule
	
    UniMatch\!\!&\hspace{-2.4mm}| \textbf{\gray{RN-101}} & \textbf{76.6} & \textbf{77.9} & \textbf{79.2} & \textbf{79.5} \\
    
    \bottomrule
    \end{tabular}
    \vspace{-2mm}
    \caption{Comparison with SOTAs on the \textbf{Cityscapes}. $\ddag$: U$^2$PL ResNet-50 results are reproduced on the same splits as ours.}
    \vspace{-2mm}
    \label{table:cityscapes}
\end{table}

%% file: table/coco.tex
\begin{table}[t]
	\centering
	\small
	
	\begin{tabular}{r|ccccc}
		\toprule
		
		\multirow{2}{*}{\textbf{COCO}} & 1/512 & 1/256 & 1/128 & 1/64 & 1/32  \\
		
		~ & (232) & (463) & (925) & (1849) & (3697)  \\ 
		
		\midrule
		
		SupBaseline & 22.9 & 28.0 & 33.6 & 37.8 & 42.2 \\
		
		PseudoSeg \cite{zou2021pseudoseg} & 29.8 & 37.1 & 39.1 & 41.8 & 43.6 \\
		
		PC$^2$Seg \cite{zhong2021pixel} & 29.9 & 37.5 & 40.1 & 43.7 & 46.1 \\
		
		\midrule
		
		UniMatch & \textbf{31.9} & \textbf{38.9} & \textbf{44.4} & \textbf{48.2} & \textbf{49.8} \\
		
		\bottomrule
	\end{tabular}
	\vspace{-2mm}
	\caption{Comparison with SOTAs on the \textbf{COCO} with Xception-65.}
	\vspace{1mm}
	\label{table:coco}
	\vspace{-4mm}
\end{table}

%% file: table/ablation_uniperb_dusperb.tex
\begin{table}[t]
	\centering
	\small
	
	\begin{tabular}{r|ccccc}
		\toprule

        Method & 92 & 183 & 366 & 732 & 1464 \\
		
		\midrule
		
		SOTA Before$^\star$ & \underline{65.2} & 71.0 & 74.6 & 77.3 & 79.1 \\
		
		FixMatch (Fig \ref{fig:fixmatch}) & 63.9 & \underline{73.0} & \underline{75.5} & \underline{77.8} & \underline{79.2} \\
		
		\midrule
		
		UniPerb (Fig \ref{fig:uniperb}) & 72.0 & 75.8 & 77.5 & 79.3 & 80.1 \\
		
		DusPerb (Fig \ref{fig:dusperb}) & 72.1 & 75.9 & 78.3 & 78.1 & 79.6 \\
		
		\midrule
		
		UniMatch (Fig \ref{fig:unimatch}) & \textbf{75.2} & \textbf{77.2} & \textbf{78.8} & \textbf{79.9} & \textbf{81.2} \\
		
		\textbf{\textcolor{bluegreen}{Gain ($\triangle$)}} & \textbf{\textcolor{bluegreen}{$\uparrow$ 11.3}} & \textbf{\textcolor{bluegreen}{$\uparrow$ 4.2}} & \textbf{\textcolor{bluegreen}{$\uparrow$ 3.3}} & \textbf{\textcolor{bluegreen}{$\uparrow$ 2.1}} & \textbf{\textcolor{bluegreen}{$\uparrow$ 2.0}} \\
		
		\bottomrule
	\end{tabular}
	\vspace{-2mm}
	\caption{Effectiveness of each component on the \textbf{Pascal}. $\star$: Borrowed from results in Table \ref{table:pascal_original}, that use the same size (321) as us. The \underline{better results} between our reproduced FixMatch and previous SOTAs are underlined. The \textcolor{bluegreen}{$\triangle$} is measured against FixMatch.}
	\vspace{-2mm}
	\label{table:ablation_uniperb_dusperb}
\end{table}

%% file: table/citysapes_unimatch_fixmatch.tex
\begin{table}[t]
\setlength\tabcolsep{3mm}
	\centering
	\small
	
	\begin{tabular}{r|cccc}
		\toprule
		
		Method & 1/16 & 1/8 & 1/4 & 1/2 \\
		
		\midrule
		
	    FixMatch (Baseline) & 72.6 & 75.7 & 76.8 & 78.2 \\
		
		UniMatch (Ours) & \textbf{75.0} & \textbf{76.8} & \textbf{77.5} & \textbf{78.6} \\
		
		\textbf{\textcolor{bluegreen}{Gain ($\triangle$)}} & \textbf{\textcolor{bluegreen}{$\uparrow$ 2.4}} & \textbf{\textcolor{bluegreen}{$\uparrow$ 1.1}} & \textbf{\textcolor{bluegreen}{$\uparrow$ 0.7}} & \textbf{\textcolor{bluegreen}{$\uparrow$ 0.4}} \\
		
		\bottomrule
	\end{tabular}
	\vspace{-2mm}
	\caption{Comparison between our UniMatch and our reproduced strong FixMatch baseline on the \textbf{Cityscapes} dataset.}
	\vspace{-2mm}
	\label{table:citys_unimatch_fixmatch}
\end{table}

%% file: table/coco_unimatch_fixmatch.tex
\begin{table}[t]
\setlength\tabcolsep{3mm}
	\centering
	\small
	
	\begin{tabular}{r|ccccc}
		\toprule
		
		Method & 1/512 & 1/256 & 1/128 & 1/64 & 1/32 \\
		
		\midrule
		
	    FixMatch & 26.8 & 32.8 & 37.7 & 44.1 & 47.5 \\
		
		UniMatch & \textbf{31.9} & \textbf{38.9} & \textbf{44.4} & \textbf{48.2} & \textbf{49.8} \\
		
		\textbf{\textcolor{bluegreen}{Gain ($\triangle$)}} & \textbf{\textcolor{bluegreen}{$\uparrow$ 5.1}} & \textbf{\textcolor{bluegreen}{$\uparrow$ 6.1}} & \textbf{\textcolor{bluegreen}{$\uparrow$ 6.7}} & \textbf{\textcolor{bluegreen}{$\uparrow$ 4.1}} & \textbf{\textcolor{bluegreen}{$\uparrow$ 2.3}} \\
		
		\bottomrule
	\end{tabular}
	\vspace{-2mm}
	\caption{Comparison between our UniMatch and our reproduced strong FixMatch baseline on the \textbf{COCO} dataset.}
	\vspace{-4mm}
	\label{table:coco_unimatch_fixmatch}
\end{table}

%% file: table/ablation_3views.tex
\begin{table}[t]
\setlength\tabcolsep{2.5mm}
	\centering
	\small
	
	\begin{tabular}{r|ccccc}
		\toprule
		
		Method & 92 & 183 & 366 & 732 & 1464 \\

		\midrule
		
		Dual Image Views & 72.1 & 75.9 & 78.3 & 78.1 & 79.6 \\
		
		Triple Image Views & 71.6 & 76.4 & 78.4 & 78.8 & 79.6 \\
		
		\midrule
		
		 UniMatch & \textbf{75.2} & \textbf{77.2} & \textbf{78.8} & \textbf{79.9} & \textbf{81.2} \\
		
		\bottomrule
	\end{tabular}
	\vspace{-2mm}
	\caption{Ablation study on the non-trivial improvement of diverse perturbations. Our UniMatch is consistently superior to its counterpart which directly uses triple strongly perturbed images as inputs.}
	\vspace{-2mm}
	\label{table:ablation_3views}
\end{table}

%% file: table/ablation_dusperb_2xbs.tex
\begin{table}[t]
\setlength\tabcolsep{3mm}
    \small
    \centering
	
	\begin{tabular}[t]{r|ccccc}
		\toprule
		Method & 92 & 183 & 366 & 732 & 1464 \\

		\midrule
		
		$2\times$ Batch Size & 62.5 & 74.5 & 77.1 & 77.8 & 79.3 \\
		
		$2\times$ Epochs & 61.8 & 73.6 & 76.2 & 77.6 & 79.4 \\
		
		\midrule
		
		DusPerb & \textbf{72.1} & \textbf{75.9} & \textbf{78.3} & \textbf{78.1} & \textbf{79.6} \\
		
		\bottomrule
	\end{tabular}
	\vspace{-2mm}
	\caption{Ablation study on the necessity of dual-stream perturbations, compared with doubling the batch size or training epochs.}
	\vspace{-4mm}
	\label{table:ablation_dusperb_2xbs}
\end{table}

%% file: table/ablation_shared_stream.tex
\begin{table}[t]
\setlength\tabcolsep{2.4mm}
    \small
    \centering
	
	\begin{tabular}[t]{r|ccccc}
		\toprule

		Method & 92 & 183 & 366 & 732 & 1464 \\

		\midrule
		
		FixMatch & 63.9 & 73.0 & 75.5 & 77.8 & 79.2 \\
		
		Single Hybrid View & 63.4 & 72.8 & 75.0 & 76.9 & 78.7 \\
		
		\midrule
		
		Dual Hybrid Views & 71.8 & 74.6 & 76.8 & 78.4 & 79.6 \\ 
		
		UniPerb & \textbf{72.0} & \textbf{75.8} & \textbf{77.5} & \textbf{79.3} & \textbf{80.1} \\
		
		\bottomrule
	\end{tabular}
	\vspace{-2mm}
	\caption{Ablation study on separating image- and feature-level perturbations into independent streams.}
	\vspace{-2mm}
	\label{table:ablation_shared_stream}
\end{table}

%% file: table/ablation_num_of_views.tex
\begin{table}[t]
	\centering
	\small
    \setlength\tabcolsep{1.6mm}
	\begin{tabular}{cc|ccccc|ccc}
		\toprule
        
		IS & FS & 92 & 183 & 366 & 732 & 1464 & 1/16 & 1/8 & 1/4 \\
		
		\midrule
		
		2 & 1 & 75.2 & 77.2 & 78.8 & \textbf{79.9} & \textbf{81.2} & 76.5 & 77.0 & 77.2 \\
		
		2 & 2 & 75.2 & \textbf{77.7} & 78.9 & 79.7 & 80.7 & \textbf{76.9} & 77.3 & \textbf{77.9} \\
		
		3 & 3 & \textbf{75.5} & 77.0 & 78.6 & 79.5 & 80.5 & 76.7 & \textbf{77.7} & 77.3 \\
		
		4 & 4 & 75.0 & 76.6 & \textbf{79.4} & 79.8 & 80.6 & 76.6 & 77.1 & 77.5 \\
		
		\bottomrule
	\end{tabular}
	\vspace{-2mm}
	\caption{The performance change with respect to the number of image- and feature-level perturbation streams. \textbf{IS} stands for image-level stream, while \textbf{FS} represents feature-level stream. The first row (IS:2, FS:1) is our UniMatch approach.}
	\vspace{-4mm}
	\label{table:ablation_num_of_views}
\end{table}

%% file: table/cd_whu_levir.tex
\begin{table*}[t]
\setlength\tabcolsep{1.4mm}
	\centering
	\small
	
    \begin{tabular}{r|cccc|cccc} 
		\toprule
		\multirow{2}{*}{Method} & \multicolumn{4}{c|}{WHU-CD} & \multicolumn{4}{c}{LEVIR-CD} \\ 
		
		\cmidrule{2-9}
		
		~ & 5\% & 10\% & 20\% & 40\% & 5\% & 10\% & 20\% & 40\% \\
		
		\midrule
		
		S4GAN \cite{mittal2019semi} & 18.3 / 96.69 & 62.6 / 98.15 & 70.8 / 98.60 & 76.4 / 98.96 & 64.0 / 97.89 & 67.0 / 98.11 & 73.4 / 98.51 & 75.4 / 98.62 \\
		
		SemiCDNet \cite{peng2020semicdnet} & 51.7 / 97.71 & 62.0 / 98.16 & 66.7 / 98.28 & 75.9 / 98.93 & 67.6 / 98.17 & 71.5 / 98.42 & 74.3 / 98.58 & 75.5 / 98.63 \\ 
		
		SemiCD \cite{bandara2022revisiting} & 65.8 / 98.37 & 68.1 / 98.47 & 74.8 / 98.84 & 77.2 / 98.96 & 72.5 / 98.47 & 75.5 / 98.63 & 76.2 / 98.68 & 77.2 / 98.72 \\
		
		\midrule

        SupBaseline & 48.3 / 97.41 & 60.7 / 98.03 & 69.7 / 98.55 & 69.5 / 98.47 & 67.5 / 98.12 & 73.4 / 98.50 & 75.2 / 98.63 & 77.7 / 98.79 \\
		
		UniMatch | \textbf{\gray{PSPNet}} & \textbf{77.5} / \textbf{99.06} & \textbf{78.9} / \textbf{99.10} & \textbf{82.9} / \textbf{99.26} & \textbf{84.4} / \textbf{99.32} & \textbf{75.6} / \textbf{98.62} & \textbf{79.0} / \textbf{98.83} & \textbf{79.0} / \textbf{98.84} & \textbf{78.2} / \textbf{98.79} \\

        \midrule

        SupBaseline & 54.1 / 97.56 & 60.9 / 97.86 & 68.4 / 98.34 & 76.2 / 98.87 & 69.3 / 98.28 & 76.0 / 98.69 & 77.6 / 98.79 & 80.5 / 98.94 \\

        UniMatch | \textbf{\gray{DeepLab}} & \textbf{80.2} / \textbf{99.15} & \textbf{81.7} / \textbf{99.22} & \textbf{81.7} / \textbf{99.18} & \textbf{85.1} / \textbf{99.35} & \textbf{80.7} / \textbf{98.95} & \textbf{82.0} / \textbf{99.02} & \textbf{81.7} / \textbf{99.02} & \textbf{82.1} / \textbf{99.03}  \\

		\bottomrule
	\end{tabular}
	\vspace{-2mm}
 
	\caption{Results on the \textbf{WHU-CD} \cite{ji2018fully} and \textbf{LEVIR-CD} \cite{chen2020spatial} datasets. Numbers in each cell denote \emph{changed-class IOU} and \emph{overall accuracy}, respectively. The fraction (\eg, 5\%) denotes the proportion of labeled images. We try both PSPNet and DeepLabv3+ with ResNet-50.}
	\vspace{-3mm}
	\label{tab:whu_levir_cd}
\end{table*}

%% file: table/ablation_fp_location.tex
\begin{table}[t]
\small
\centering

\begin{tabular}[t]{r|ccccc}
		\toprule
		Location & 92 & 183 & 366 & 732 & 1464 \\

		\midrule
		
		Decoder-Classifier & 64.5 & 72.2 & 76.9 & 77.9 & 79.5 \\

        En-Decoder (Default) & \textbf{72.0} & \textbf{75.8} & \textbf{77.5} & \textbf{79.3} & \textbf{80.1} \\
		
		\bottomrule
	\end{tabular}
	\vspace{-2mm}
	\caption{Ablation study on the location to insert feature perturbations in our UniPerb method.}
	\vspace{-4mm}
	\label{tab:ablation_fp_location}
\end{table}

%% file: table/acdc.tex
\begin{table}[t]
\setlength\tabcolsep{3.5mm}
    \centering
    \small

    \begin{tabular}{r|cccc}
	\toprule
	Method & 1 case & 3 cases & 7 cases \\

	\midrule

        SupBaseline & 28.5 & 41.5 & 62.5 \\
        
	UA-MT \cite{yu2019uncertainty} & N/A & 61.0 & 81.5 \\
    
	CPS \cite{chen2021semi} & N/A & 60.3 & 83.3 \\
		
	CNN \& Trans \cite{luo2021semi} & N/A & 65.6 & 86.4 \\
	
	\midrule
	
	UniMatch (Ours) & \textbf{85.4} & \textbf{88.9} & \textbf{89.9} \\
	
	\bottomrule
    \end{tabular}
    \vspace{-2mm}
    \caption{Comparison with SOTAs on \textbf{ACDC} \cite{bernard2018deep} with 1/3/7 labeled cases. There are 70 training cases in total. Results are measured by Dice Similarity Coefficient (DSC) metric averaged on 3 classes.}
    \vspace{-5mm}
    \label{tab:acdc}
\end{table}

%% file: section/conclusion.tex
\section{Conclusion}

We investigate the promising role of FixMatch in semi-supervised semantic segmentation. We first present that equipped with proper image-level strong perturbations, a vanilla FixMatch can indeed outperform the SOTAs. Inspired by this, we further strengthen its perturbation practice from two perspectives. On one hand, we unify image- and feature-level perturbations to form a more diverse perturbation space. On the other, we design a dual-stream perturbation technique to fully exploit image-level perturbations. Both components facilitate our baseline significantly. The final method UniMatch improves previous results remarkably in all the natural, medical, and remote sensing scenarios.

\vspace{4mm}
\noindent
\textbf{Acknowledgements.} This work is supported by the NSFC Program (62222604, 62206052, 62192783), CAAI-Huawei MindSpore (CAAIXSJLJJ-2021-042A), China Postdoctoral Science Foundation Project
(2021M690609), Jiangsu Natural Science Foundation Project (BK20210224), and CCF-Lenovo Bule Ocean Research Fund.

We are sincerely grateful to Boyao Shi (NJU) for polishing the figures, and Zhen Zhao (USYD) for comments.

We also thank the support from SenseTime Research.

%% file: section/supp_singlecolumn.tex
\appendix

\begin{table*}[t]
	\centering
	\small
	
	\begin{tabular}{l|ccccc|ccc}
		\toprule

		Perturbation Level & 92 & 183 & 366 & 732 & 1464 & 1/16 & 1/8 & 1/4 \\

		\midrule
		\midrule
		
		Image Level Alone (FixMatch) & 63.9 & 73.0 & 75.5 & 77.8 & 79.2 & 74.1 & 75.9 & 76.4 \\
		
		Feature Level Alone & 66.0 & 69.6 & 74.0 & 77.3 & 78.9 & 73.7 & 74.5 & 76.4 \\
		
		\midrule
		
		Unified Levels (UniPerb) & 72.0 & 75.8 & 77.5 & 79.3 & 80.1 & 76.0 & 76.9 & 76.6 \\
		
		\bottomrule
	\end{tabular}
	\caption{Results (\%) of only using \emph{single} perturbation level, either image-level perturbations (original FixMatch) or feature-level perturbations. These results are obtained from the Pascal dataset with DeepLabv3+ and ResNet-101. We also provide results of our UniPerb (Unified Levels) as a reference, which unifies the two different levels of perturbations.}
	\label{table:ablation_fp_only}
\end{table*}

\begin{table*}[t]
	\centering
	\small
	
	\vspace{1mm}
	
	\begin{tabular}{l|ccccc|ccc}
		\toprule

		Method & 92 & 183 & 366 & 732 & 1464 & 1/16 & 1/8 & 1/4 \\

		\midrule
		\midrule
		
		Single-Stream FP (UniPerb) & 72.0 & 75.8 & 77.5 & 79.3 & 80.1 & 76.0 & 76.9 & 76.6 \\
		
		Dual-Stream FP & \textbf{73.4} & \textbf{77.1} & \textbf{78.5} & \textbf{79.6} & \textbf{80.4} & \textbf{76.2} & \textbf{77.0} & \textbf{76.8} \\
		
		\bottomrule
	\end{tabular}
	\caption{Effectiveness of dual-stream perturbations at the feature level (\%). FP here denotes feature perturbation. Same as the single-stream FP (UniPerb), the dual-stream FP also contains one stream for image-level strong perturbations.}
	\label{table:ablation_uniperb_dual}
\end{table*}

\begin{table*}[t]
	\centering
	\small
	
	\begin{tabular}{l|ccccccc}
		\toprule

        \multirow{2}{*}{Number of labeled images} & \multicolumn{7}{c}{Number of image-level perturbation streams} \\
        
        \cmidrule{2-8}
        
		~ & 1 (FixMatch) & 2 (DusPerb) & 3 & 4 & 5 & 6 & 7 \\

		\midrule
		\midrule
		
		High-quality set: 732 & 77.8 & 78.1 & 78.8 & 78.9 & \textbf{79.1} & 78.7 & 78.4 \\
		
		Blended set: 662 (1/16) & 74.1 & 75.3 & 75.4 & 76.1 & 76.0 & \textbf{76.7} & 76.3 \\

		\bottomrule
	\end{tabular}
	\caption{The performance (\%) change with respect to the number of image-level strong perturbation streams.}
	\label{table:ablation_num_of_strong}
\end{table*}

\section{How about Removing Image-Level Strong Perturbations?}

It has almost become a primary concern and a common practice in various semi-supervised settings to seek proper image-level strong perturbations first. Nevertheless, this process requires many time-consuming trials and delicate selection of different combinations. To make matters worse, in some domain-specific tasks, such as medical image analysis, it is challenging for most practitioners to figure out appropriate ones. Therefore, a natural question raises: could we replace image-level strong perturbations in FixMatch with a simple channel dropout perturbation at the feature level?

To validate this, we make a modification to original FixMatch that, input images are not processed by any strong data augmentation, but their features are perturbed by a channel dropout. It can be observed from Table~\ref{table:ablation_fp_only} that in most cases, feature-level perturbation alone can indeed perform on par with original image-level strong perturbations in FixMatch, merely slightly inferior. Hence, we believe that such simple but universal feature perturbations may serve as a promising supplement, when image-level strong perturbations fail to work in some rarely explored scenarios.

\section{Dual-Stream Feature-Level Perturbations}

The technique of dual-stream perturbations has been proved to be highly beneficial at the image level. Certainly, we wish to check its effectiveness at the feature level. Thus, we attempt to strengthen our proposed UniPerb via performing twice parallel channel dropout on the extracted features. The dual perturbed features are then sent into the decoder to produce two final predictions for learning. The results of dual-stream feature-level perturbations are reported in Table~\ref{table:ablation_uniperb_dual}. Our UniPerb can be further boosted via maintaining dual feature perturbation streams. As discussed in Section~3.3 of our main paper, we conjecture that dual random perturbations on the same features can also be considered to produce a pair of positive views, thereby harvesting the merits of contrastive learning. Despite the effectiveness, we decide not to conduct dual-stream feature perturbations in our main approach, because the current version is powerful enough, and we hope to avoid additional computational burden during training.

\section{More Image-Level Perturbation Streams}

Here, the auxiliary feature-level stream is excluded, which means the perturbation space is completely constrained at the image level. Then, we progressively increase the number of image-level strong perturbation streams on the Pascal, and report the performance in Table \ref{table:ablation_num_of_strong}.

The performance is steadily improved as the number of strong views is increased to a certain number. But if we continue to increase beyond it, then the performance might drop a little. The results indicate that, two or three strong views are already enough to fully probe the original image-level perturbation space. Excessive strong views might cause the model to struggle in learning every single view.

\section{Limitations, Discussions, and Future Works}

In our UniMatch, a confidence threshold is primarily set to suppress potentially incorrect pseudo labels. In some challenging scenarios, \eg COCO, however, we observe that around 15\% pixels are discarded during the learning course. Therefore, how to make full use of these uncertain pixels and meantime avoid error accumulation will be a promising direction to further facilitate current semi-supervised algorithms. This may also enable our model to be more robust to different thresholds.

Moreover, our framework, along with its precedents, such as the FixMatch serials \cite{sohn2020fixmatch, berthelot2019mixmatch, berthelot2019remixmatch, zhang2021flexmatch} and UDA \cite{xie2019unsupervised}, heavily relies on the pseudo labeling quality on unlabeled images. In case yielded pseudo labels are poor, it would be hard for our semi-supervised learner to mine meaningful knowledge from unlabeled images. Therefore, if the class distribution is highly imbalanced, the model will be gradually biased to majority classes during training and pseudo labeling, making the minority classes worse and worse. In addition, on common benchmarks, the domain gap between labeled and unlabeled images is rarely considered. However, in real worlds, the abundant unlabeled images can not share exactly the same domain as labeled ones. The semi-supervised learner could benefit from more unlabeled images if domain shift is well addressed.

Last but not least, existing academic settings in semi-supervised classification/segmentation/detection prefer to restricting labeled images to an extremely low proportion, \eg only providing 40 labels on the CIFAR-10 and 92 labeled images on the Pascal. Nevertheless, considering most real-world demands, it might be more practical to assume labeled images is in the tens of thousands, while unlabeled images are even more, might in millions. Actually, a prior work \cite{zoph2020rethinking} already explored such a setting, but it is expected to be further improved, both in accuracy and training efficiency.

We leave the aforementioned four problems, namely 1) how to fully exploit uncertain pixels, 2) class imbalance in pseudo labeling, 3) domain shift in pseudo labeling, and 4) how to effectively benefit from millions of unlabeled samples together with considerable labeled ones, to our future works.